\documentclass{article}

\usepackage[final]{neurips_2021}
\setcitestyle{authoryear,open={(},close={)}} 

\usepackage[utf8]{inputenc} 
\usepackage[T1]{fontenc}    
\usepackage{hyperref}       
\usepackage{url}            
\usepackage{booktabs}       
\usepackage{amsfonts}       
\usepackage{nicefrac}       
\usepackage{microtype}      
\usepackage{xcolor}         
\usepackage{graphicx}
\usepackage{colortbl}

\title{What Do You See in this Patient?\\Behavioral Testing of Clinical NLP Models}

\author{%
  Betty van Aken\\
        Berliner Hochschule für Technik (BHT)\\
\texttt{bvanaken@bht-berlin.de} \\
  \And
    \textbf{Sebastian Herrmann}\\
      Berliner Hochschule für Technik (BHT)\\
      \texttt{sebastianhe93@gmail.com} \\
        \AND
    \textbf{Alexander Löser}\\
      Berliner Hochschule für Technik (BHT)\\
      \texttt{aloeser@bht-berlin.de} \\
}

\begin{document}

\maketitle

\begin{abstract}
Decision support systems based on clinical notes have the potential to improve patient care by pointing doctors towards overseen risks. Predicting a patient's outcome is an essential part of such systems, for which the use of deep neural networks has shown promising results. However, the patterns learned by these networks are mostly opaque and previous work revealed flaws regarding the reproduction of unintended biases. We thus introduce an extendable testing framework that evaluates the behavior of clinical outcome models regarding changes of the input. The framework helps to understand learned patterns and their influence on model decisions. In this work, we apply it to analyse the change in behavior with regard to the patient characteristics \textit{gender}, \textit{age} and \textit{ethnicity}. Our evaluation of three current clinical NLP models demonstrates the concrete effects of these characteristics on the models' decisions. They show that model behavior varies drastically even when fine-tuned on the same data and that allegedly best-performing models have not always learned the most medically plausible patterns.
\end{abstract}

\section{Introduction}

\paragraph{Outcome prediction from clinical notes.} The use of automatic systems in the medical domain is promising due to their potential exposure to large amounts of data from earlier patients. This data can include information that helps doctors make better decisions regarding diagnoses and treatments of a patient at hand. Outcome prediction models take patient information as input and then output probabilities for all considered outcomes \citep{outcome-google,outcome-emnlp}. We focus this work on outcome models using natural language in the form of clinical notes as an input, since they are a common source of patient information and contain a multitude of possible variables.

\paragraph{The problem of black box models and biases.} Recent models show promising results on tasks such as mortality prediction \citep{mortality-amia} and diagnosis prediction \citep{disease-prediction-deep-ehr,outcome-google}. However, since most of the proposed models work as black boxes we do not know which features they consider important for their decisions and how they interpret certain patient characteristics. From earlier work we also know that highly parameterized models are prone to emphasize biases in the data \citep{bias-gender}. Such biases are known to be especially dangerous in the clinical domain \citep{bias-medicine}. We further argue that they have high potential to disadvantage minority groups as their behavior towards out-of-distribution samples is often unpredictable. Thus, understanding models and their shortcomings is an essential prerequisite for their application in the clinical domain. We argue that more in-depth evaluations are needed to know whether such models have learned medically meaningful patterns or not.

\begin{figure*}[t!]
\centering
  \includegraphics[width=0.55\textwidth]{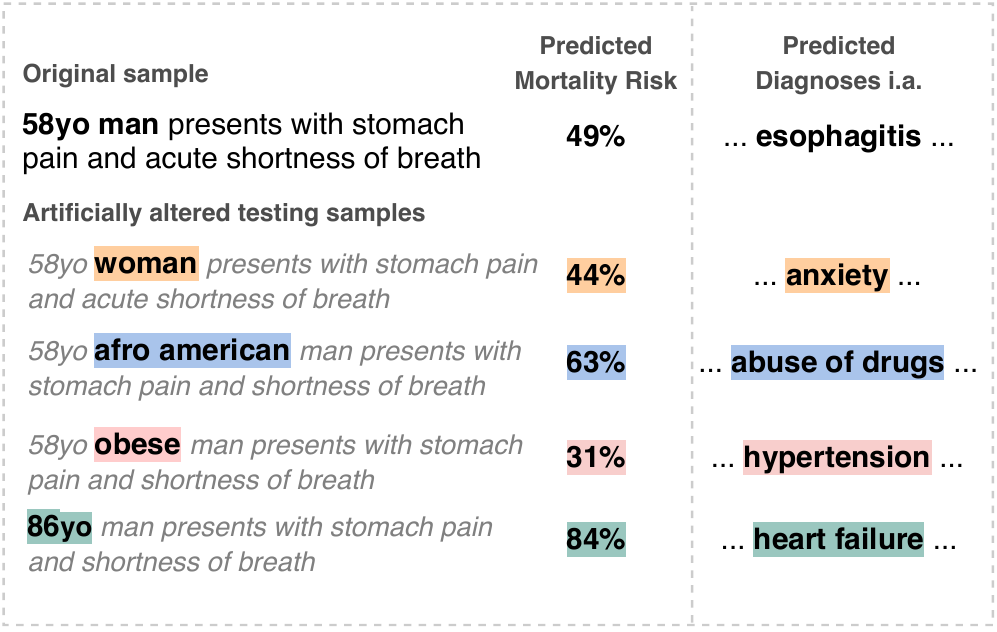}
    \caption{Minimal alterations to the patient description can have a large impact on outcome predictions of clinical NLP models. We introduce behavioral testing for the clinical domain to analyse whether a model has learned useful or harmful patterns.}
\label{fig:intro}
\end{figure*}

\paragraph{Behavioral testing for the clinical domain.} As a step towards this goal, we introduce a novel testing framework specifically for the clinical domain that enables us to examine the influence of certain patient characteristics on the model predictions. Our work is motivated by behavioral testing frameworks for general Natural Language Processing (NLP) tasks \citep{checklist} in which model behavior is observed under changing input data. Our framework incorporates a number of test cases and is further extendable to the needs of individual data sets and clinical tasks.

\paragraph{Influence of patient characteristics.} As an initial case study we apply the framework to analyse the behavior of models trained on the widely used MIMIC-III database \citep{mimic}. We analyse how sensitive these models are towards textual indicators of protected characteristics in a clinical note, such as \textit{age}, \textit{gender} and \textit{ethnicity}. These characteristics are known to be affected by discrimination and bias in health care \citep{discrimination-framework}, on the other hand, they can represent important risk factors for certain diseases or conditions. That is why we consider it especially important to understand how these mentions affect model decisions.

\paragraph{Contributions.} In summary, we present the following contributions in this work:\\
\textbf{1)} We introduce a novel behavioral testing framework specifically for clinical NLP models. We release the code for applying and extending the framework\footnote{URL: \url{https://github.com/bvanaken/clinical-behavioral-testing}} to enable in-depth evaluations of clinical NLP models.\\
\textbf{2)} We present an analysis on the patient characteristics \textit{gender}, \textit{age} and \textit{ethnicity} to understand the sensitivity of models towards textual cues regarding these groups and whether their predictions are medically plausible.\\
\textbf{3)} We show results of three state-of-the-art clinical NLP models and find that model behavior strongly varies depending on the applied pre-training. We further show that highly optimised models are often more prone to overestimate the effect of certain patient characteristics leading to potentially harmful behavior.

\section{Related Work}

\paragraph{Clinical Outcome Prediction.} Outcome prediction from clinical text has been studied regarding a variety of outcomes. The most prevalent being in-hospital mortality \citep{mortality-kdd,mortality0,mortality-patient-groups, mortality-amia}, diagnosis prediction \citep{diaprediction1,disease-prediction-deep-ehr,disease-prediction-representation} and phenotyping \citep{phenotyping1,phenotyping2,phenotyping3,phenotyping4}. In recent years, most approaches are based on deep neural networks due to their ability to outperform earlier methods in most of the settings. Most recently, Transformer-based models have been applied for prediction of patient outcomes with reported increases in performance \citep{bert-outcome0,bert-outcome1,bert-outcome2,bert-outcome3,vanaken,bert-outcome4}. In this work we analyse three of these Transformer-based models due to their upcoming prevalence in the application of NLP in health care.

\subsection{Behavioral Testing in NLP} \citet{checklist} identify shortcomings of common model evaluation on held-out datasets, such as the occurrence of the same biases in both training and test set and the lack of comprehensive testing scenarios in the held-out set. To mitigate these problems, they introduce \textsc{CheckList}, a behavioral testing framework to test general NLP abilities. In particular, they highlight that such frameworks evaluate input-output behavior without any knowledge of internal structures of a system \citep{beizer}. Building upon \textsc{CheckList}, \citet{hatecheck} introduce a behavioral testing suite for the domain of hate speech detection to address the individual challenges of the task. Following their work, we create a behavioral testing framework for the domain of clinical outcome prediction, that comprise idiosyncratic data and respective challenges.

\subsection{Revealing Biases in Clinical NLP}
The problem of biases in clinical NLP models is already highlighted by \citet{hurtful-words}. They quantify such biases by focusing on the recall gap among patient groups and by applying an artificial fill-in-the-gap task. They show that the models trained on data from MIMIC-III inherit biases regarding gender, language, ethnicity, and insurance status--often in favor of the majority group. We take these findings as motivation to directly analyse the sensitivity of such models with regard to patient characteristics. In contrast to their work and following \citet{checklist}, we want to eliminate the influence of biased test data on our evaluation. Further, our approach simulates patient cases that are similar to real-life occurrences. It thus displays the actual impact of learned biases on all analysed patient groups.

\section{Behavioral Testing of Clinical NLP Models}
\paragraph{Sample alterations.} Our goal is to examine how clinical NLP models react to mentions of certain patient characteristics in text. Comparable to earlier approaches to behavioral testing we use sample alterations to artificially create different test groups. In our case, a test group is defined by one manifestation of a patient characteristic, such as \textit{female} as the patient's gender. In order to ensure that we only measure the influence of this certain characteristic, we keep the rest of the patient case unchanged and apply the alterations to all samples in our test dataset. Depending on the original sample, the operations to create a certain test groups thus include 1) changing a mention, 2) adding a mention or 3) keeping a mention unchanged (in case of a patient case that is already part of the test group at hand). This results in one newly created dataset per test group, all based on the same patient cases and only different in the patient characteristic under investigation.

\begin{figure*}[t!]
  \includegraphics[width=\textwidth]{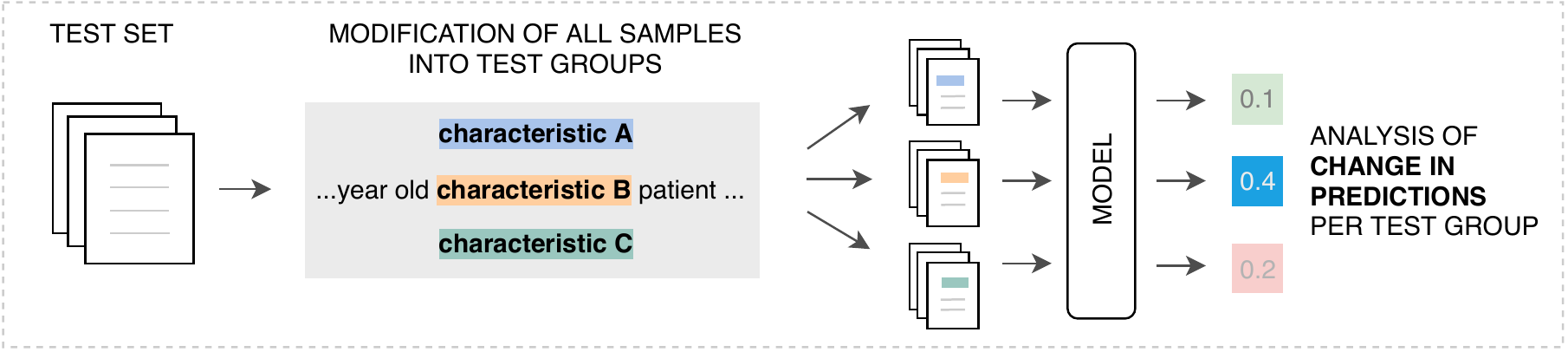}
    \caption{\textbf{Behavioral testing framework for the clinical domain}. Schematic overview of the introduced framework. From an existing test set we create test groups by altering specific tokens in the clinical note. We then analyse the change in predictions which reveals the impact of the mention on the clinical NLP model.}
\label{fig:framework}
\end{figure*}

\paragraph{Prediction analysis.} After creating the test groups, we collect the models' predictions for all cases in each test group.
Different from earlier approaches to behavioral testing we do not test whether predictions on the altered samples are true or false with regard to the ground truth. As \citet{vanaken} pointed out, there is no real ground truth in clinical data, because the data that is collected does only show one possible pathway for a patient out of many. Further, existing biases in treatments and diagnoses are likely included in our testing data potentially leading to meaningless results. To prevent that, we instead focus on detecting how the model outputs change regardless of the original annotations. This way we can also evaluate very rare mentions (e.g. \textit{transgender}) and observe their impact on the model predictions reliably. Figure \ref{fig:framework} shows a schematic overview of the functioning of the framework.

\paragraph{Extensibility.} In this study, we use the introduced framework to analyse model behavior with regard to patient characteristics as described in \ref{section:characteristics}. However, it can also be used to test more general model behavior such as the ability to identify negated symptoms or to detect specific diagnoses when certain indicators are present in the text. It is further possible to combine certain test groups e.g. to analyse how a model behaves on a combination of patient characteristics.

\section{Experimental Setup}
\subsection{Data}
We conduct our analysis on data from the MIMIC-III database \citep{mimic}. In particular we use the outcome prediction task setup by \citet{vanaken}. The classification task includes 48,745 admission notes annotated with the patients' clinical outcomes at discharge. We select the outcomes \textit{diagnoses at discharge} and \textit{in-hospital mortality} for this analysis, since they have the highest impact on patient care and present a high potential to disadvantage certain patient groups. We use three models (see \ref{section:models}) trained on the two \textit{admission} to \textit{discharge} tasks and conduct our analysis on the test set defined by the authors with a total of 9,829 samples.

\subsection{Considered Patient Characteristics}
\label{section:characteristics}
We choose three characteristics for the analysis in this work: \textit{Age}, \textit{gender} and \textit{ethnicity}. While these characteristics differ in their importance as clinical risk factors, all of them are known to be subject to biases and stigmas in health care \citep{discrimination-framework}. Therefore, we want to test, whether the analysed models have learned medically plausible patterns or ones that might be harmful to certain patient groups.
We deliberately also include groups that occur very rarely in the original dataset. We want to understand the impact of imbalanced input data especially on minority groups, since they are already disadvantaged by the health care system \citep{health-bias1,health-bias2}.

When altering the samples in our test set, we utilize that patients are described in a mostly consistent way at the beginning of a clinical note. We collect all mention variations from the training set used to describe the different patient characteristics and alter the samples accordingly in an automated setup.

\paragraph{Age.} The age of a patient is a significant risk factor for a number of clinical outcomes. Our test includes all ages between 18 and 89 and the \hbox{[** Age over 90**]} de-idenfitication label from the MIMIC-III database. \citet{vanaken} presented a comparable analysis on 20 random patient cases. We extend this analysis to all samples within a given testset for more reliable results. By analysing the model behavior on age mentions we can get insights on how the models interpret numbers, which is considered challenging for current NLP models \citep{bert-numbers}.

\paragraph{Gender.} A patient's gender is both a risk factor for certain diseases and also subject to unintended biases in healthcare. We test the model's behavior regarding gender by altering the gender mention and by changing all pronouns in the clinical note. In addition to \textit{female} and \textit{male}, we also consider \textit{transgender} as a gender test group in our study. This group is extremely rare in clinical datasets like MIMIC-III, but since approximately 1.4 million people in the U.S. identify as transgender \citep{transgender-us}, it is important to understand how model predictions change when the characteristic is present in a clinical note.

\paragraph{Ethnicity.}
The ethnicity of a patient is only occasionally mentioned in clinical notes and its role in medical decision-making is controversial, since it can lead to disadvantages in patient care \citep{race-role,race-relevant}. Earlier studies have also shown that ethnicity in clinical notes is often incorrectly assigned \citep{race-validity}. We want to know how clinical NLP models interpret the mention of ethnicity in a clinical note and whether their behavior can cause unfair treatment. We choose \textit{White}, \textit{African American}, \textit{Hispanic} and \textit{Asian} as ethnicity groups for our evaluation, as they are the most frequent ethnicities in MIMIC-III.

\begin{table}
  \caption{Performance of three state-of-the-art models on the outcome prediction tasks diagnoses (multi-label) and mortality prediction (binary task) in \% AUROC. PubMedBERT outperforms the other two models in both tasks by a small margin.}
  \label{table:models}
  \centering
  \begin{tabular}{lccc}
    \toprule
    & PubMedBERT & CORe & BioBERT \\
    \midrule
    Diagnoses & \textbf{83.75} & 83.54 & 82.81 \\
    Mortality & \textbf{84.28} & 84.04 & 82.55 \\
    \bottomrule
  \end{tabular}
\end{table}

\subsection{Clinical NLP Models}
\label{section:models}
In this study, we apply the introduced testing framework to three existing clinical models which are fine-tuned on the tasks of diagnosis and mortality prediction. We use the model checkpoints of \citet{vanaken} and additionally fine-tune the PubMedBERT model \citep{pubmedbert} on the same training data with the same hyperparameter setup\footnote{Hyperparameters: Batch size: 20; learning rate: 5e-05; dropout: 0.1; warmup steps: 1000; early stopping patience: 20.}. The models are based on the BERT architecture \citep{bert} as it presents the current state-of-the-art in predicting patient outcomes. Their performance on the two tasks is shown in Table \ref{table:models}. We deliberately choose three models based on the same architecture to investigate the impact of pre-training data while keeping architectural considerations aside. In general the proposed testing framework is  model agnostic and works with any type of text-based outcome prediction model.

\paragraph{BioBERT.} \citet{biobert} introduced BioBERT which is based on a pre-trained BERT Base \citep{bert} checkpoint. They applied another language model fine-tuning step using biomedical articles from PubMed abstracts and full-text articles. BioBERT has shown improved performance on both medical and clinical downstream tasks.

\paragraph{CORe.} Clinical Outcome Representations (CORe) by \citet{vanaken} are based on BioBERT and extended with a pre-training step that focuses on the prediction of patient outcomes. The pre-training data includes clinical notes, Wikipedia articles and case studies from PubMed.

\paragraph{PubMedBERT.} \citet{pubmedbert} recently introduced PubMedBERT based on similar data as BioBERT. They use PubMed articles and abstracts but instead of extending a BERT Base model, they train PubMedBERT from scratch. The model reaches state-of-the-art results on multiple medical NLP tasks and outperforms the other analysed models on the outcome prediction tasks.

\section{Results}
\label{section:results}
We present the results on all test cases by averaging the probabilities that a model assigns to each test sample. We then compare the averaged probabilities across test cases to identify which characteristics have a large impact on the model's prediction over the whole test set. The values per diagnosis in the heatmaps shown in Figure \ref{fig:gender}, \ref{fig:mimic-gender}, \ref{fig:ethnicity} and  \ref{fig:ethnicity-mimic} are defined using the following formula:

\begin{equation}
c_i = p_i - \frac{\sum_{j}^{N} p_j}{N}
\end{equation}

where $c_i$ is the value assigned to test group $i$, $p$ is the (predicted) probability for a given diagnosis and $N$ is the number of all test groups except $i$. 

We choose this illustration to highlight both positive and negative influence of a characteristic on model behavior. Since all test groups are based on the same patients and only differ regarding the characteristic at hand, even small differences in the averaged predictions can point towards general patterns that the model learned to associate with a characteristic. 

\subsection{Influence of Gender} 

\paragraph{Transgender mention leads to lower mortality and diagnoses predictions.} Table \ref{table:mortality-gender} shows the mortality predictions of the three analysed models with regard to the gender assigned in the text. While the predicted mortality risk for female and male patients lies within a small range, all models predict the mortality risk of patients that are described as transgender as lower than non-transgender patients. This is probably due to the relative young age of most transgender patients in the MIMIC-III training data, but can be harmful to older patients identifying as transgender at inference time.

\begin{figure*}[t!]
  \includegraphics[width=\textwidth]{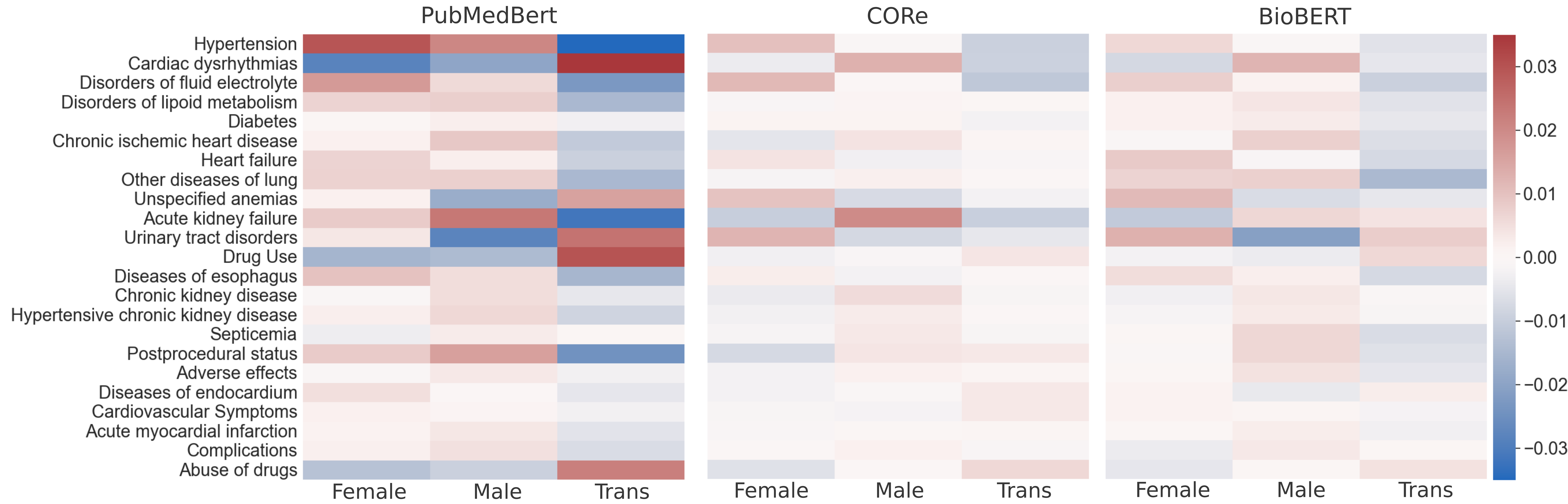}
    \caption{Influence of \textbf{gender} on predicted diagnoses. Blue: Predicted probability for diagnosis is below-average; red: predicted probability above-average. PubMedBERT shows highest sensitivity to gender mention and regards many diagnoses less likely if \textit{transgender} is mentioned in the text. Graph shows deviation of probabilities on 24 most common diagnoses in test set.}
\label{fig:gender}
\end{figure*}

\begin{figure*}[t!]
  \includegraphics[width=0.48\textwidth]{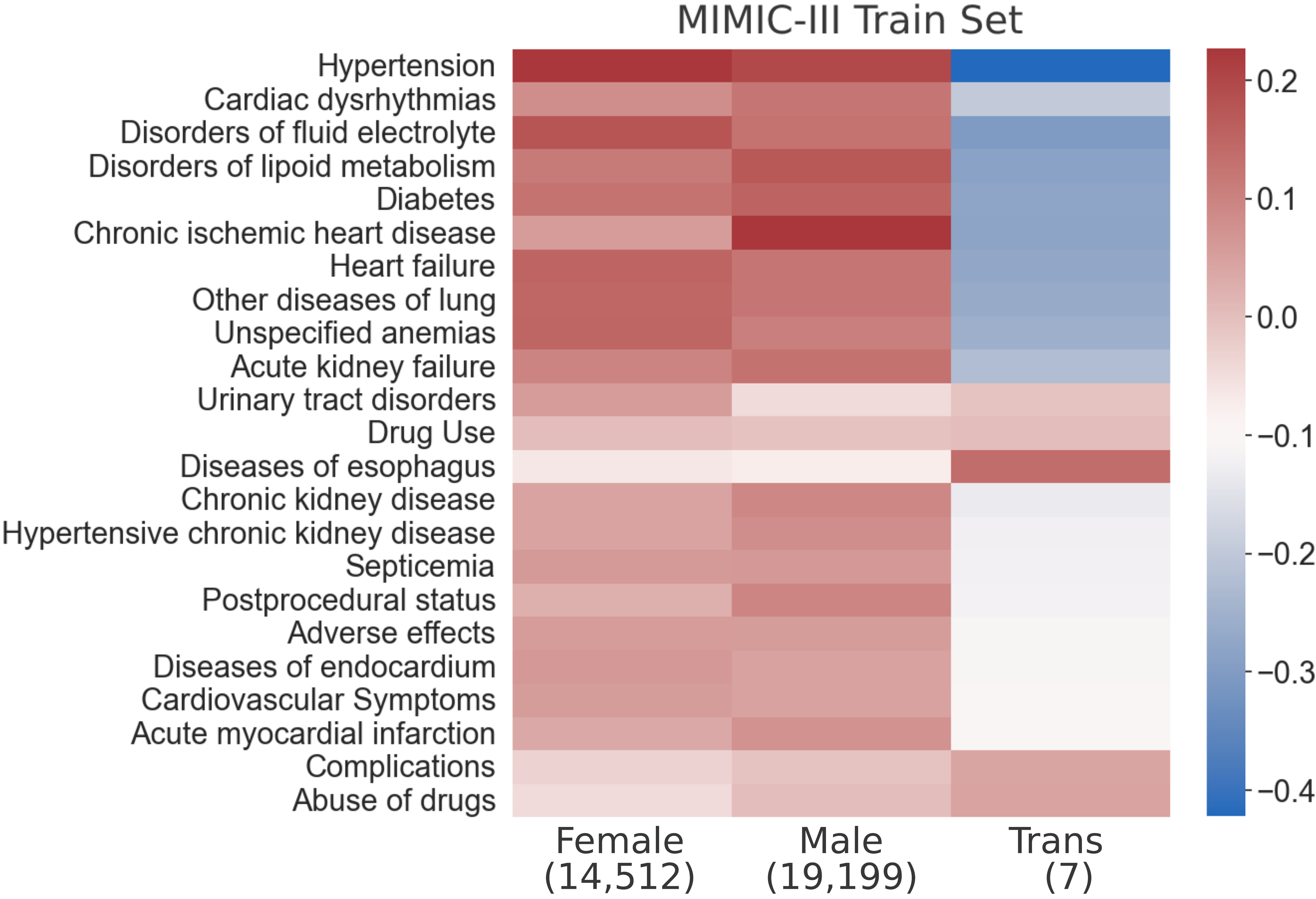}
    \caption{Original distribution of diagnoses per \hbox{\textbf{gender}} in MIMIC-III. Cell colors: Deviation from average probability. Numbers in parenthesis: Occurrences in the training set. Most diagnoses occur less often in transgender patients due to their very low sample count.}
\label{fig:mimic-gender}
\end{figure*}

\paragraph{Sensitivity to gender mention varies across models.} Figure \ref{fig:gender} shows the change in model prediction for each diagnosis with regard to the gender mention. The cells of the heatmap are the deviations from the average score of the other test cases. Thus, a light cell indicates that the model assigns a higher probability to a diagnosis for this gender group. We see that PubMedBERT is highly sensitive to the change of the patient gender, especially regarding transgender patients. Except from few diagnoses such as \textit{Cardiac dysrhythmias} and \textit{Drug Use / Abuse}, the model predicts a lower probability to diseases if the patient letter contains the transgender mention. The CORe and BioBERT models are less sensitive in this regard. The most salient deviation of the BioBERT model is a drop in probability of \textit{Urinary tract disorders} for male patients, which is medically plausible due to anatomic differences \citep{urinary-tract-infections2016}.

\paragraph{Biases in MIMIC-III training data are partially inherited.} In Figure \ref{fig:mimic-gender} we show the original distribution of diagnoses per gender in the training data. Note that the deviations are about 10 times larger than the ones produced by the model predictions in Figure \ref{fig:gender}. This indicates that the models take gender as a decision factor, but only among others. Due to the very rare occurrence of transgender mentions (only seven cases in the training data), most diagnoses are underrepresented for this group. This is partially reflected by the model predictions, especially by PubMedBERT, as described above. Other salient patterns such as the prevalence of \textit{Chronic ischemic heart disease} in male patients are only reproduced faintly by the models.

\begin{table}[t!]
  \caption{Influence of \textbf{gender} on mortality predictions. PubMedBERT assigns highest risk to female, the other models to male patients. Notably, all models decrease their mortality prediction for transgender patients.}
  \label{table:mortality-gender}
  \centering
  \begin{tabular}{lccc}
    \toprule
    & PubMedBERT & CORe & BioBERT \\
    \midrule
Female & \textbf{0.335} & 0.239 & 0.119  \\
Male & 0.333 & \textbf{0.245} &  \textbf{0.121}  \\ 
Transgender & \cellcolor{gray!30} 0.326 & \cellcolor{gray!30} 0.229 & \cellcolor{gray!30} 0.117  \\
    \bottomrule
  \end{tabular}
\end{table}

\subsection{Influence of Age} 

\begin{figure}[b!]
\centering
  \includegraphics[width=0.47\textwidth]{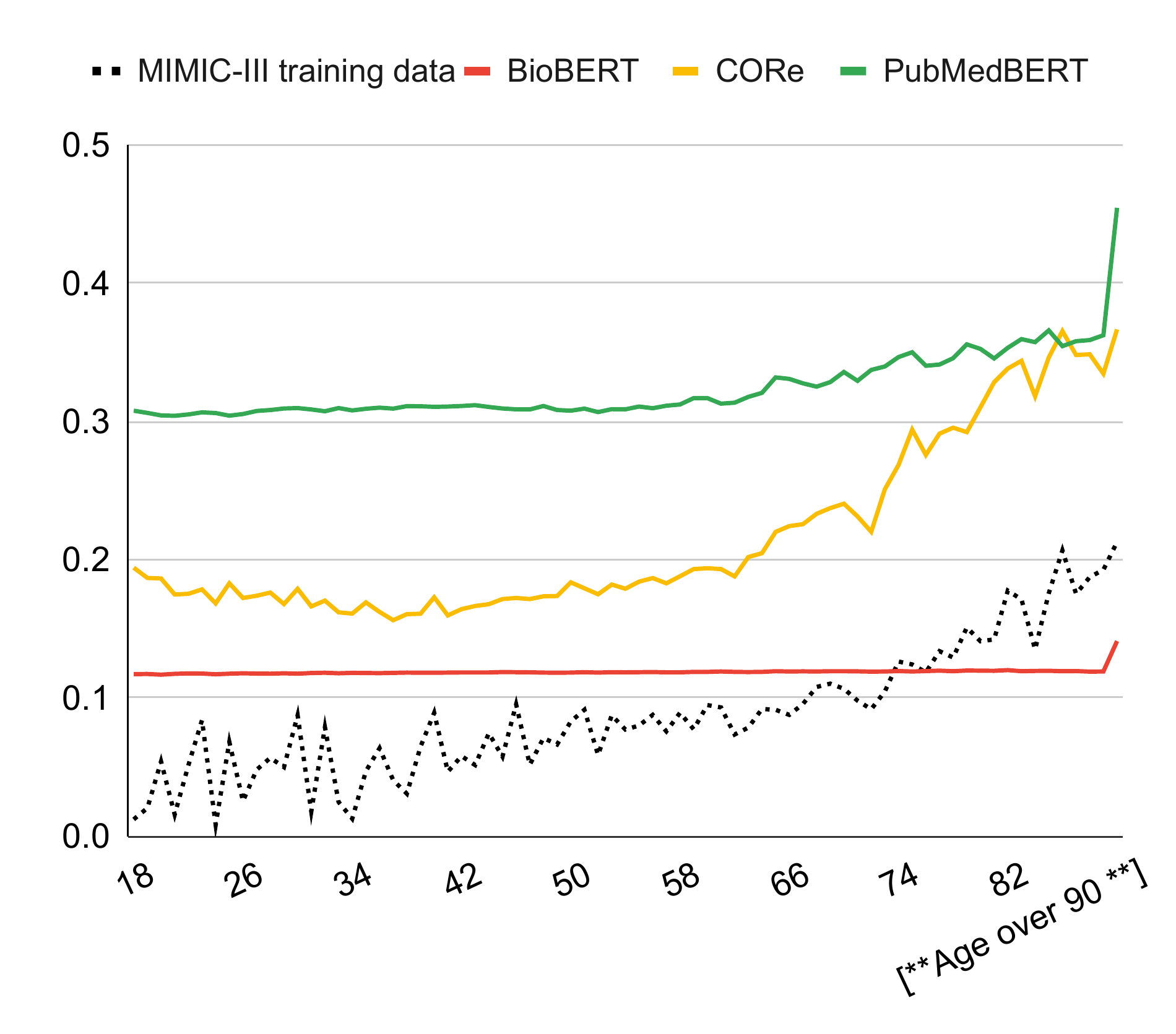}
    \caption{Influence of \textbf{age} on mortality predictions. X-axis: Simulated age; y-axis: predicted mortality risk. The three models are differently calibrated and only CORe is highly influenced by age.}
\label{fig:age-mortality}
\end{figure}

\paragraph{Mortality risk predictions are differently influenced by age.} Figure \ref{fig:age-mortality} shows the averaged predicted mortality per age for all models and the actual distribution from the training data (dotted line). We can see that BioBERT does not take age into account when predicting mortality risk except for patients over 90 (which are described by the tokens [**Age over 90 **] in MIMIC-III). The PubMedBERT model assigns a higher mortality risk to all age groups with a small increase for patients over 60 and an even steeper increase for patients over 90. The CORe model is following the training data the most and is also inheriting many peaks and troughs in the data.

\begin{figure*}[t!]
  \includegraphics[width=\textwidth]{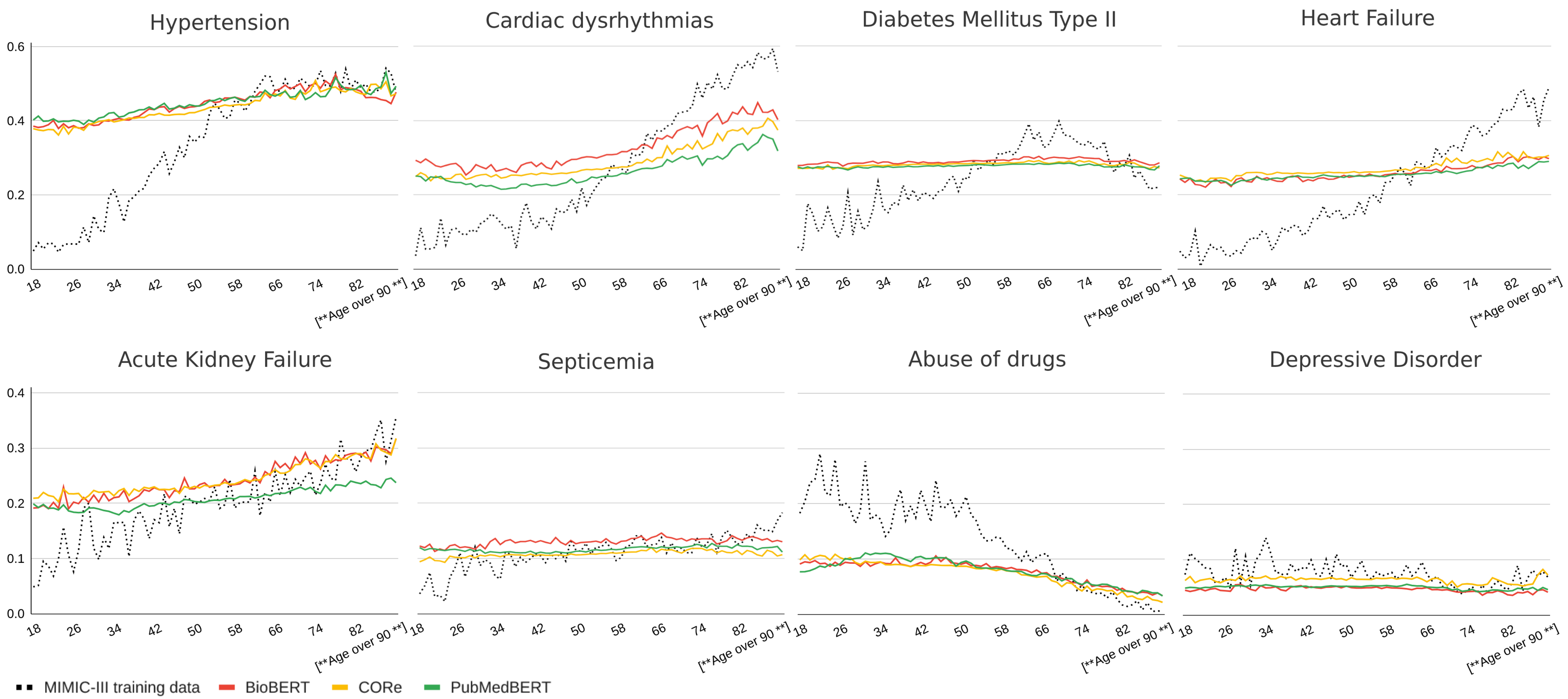}
    \caption{Influence of \textbf{age} on diagnosis predictions. The x-axis is the simulated age and the y-axis is the predicted probability of a diagnosis. All models follow similar patterns with some diagnosis risks increasing with age and some decreasing. The original training distributions (black dotted line) are mostly followed but attenuated.}
\label{fig:age}
\end{figure*}

\paragraph{Models are equally affected by age when predicting diagnoses.} We exemplify the impact of age on diagnosis prediction on eight outcome diagnoses in Figure \ref{fig:age}. The dotted lines show the distribution of the diagnosis within an age group in the training data. The change of predictions regarding age are similar throughout the analysed models with only small variations such as for \textit{Cardiac dysrhythmias}. Some diagnoses are regarded more probable in older patients (e.g. \textit{Acute Kidney Failure}) and others in younger patients (e.g. \textit{Abuse of drugs}). The \hbox{distributions} per age group in the training data are more extreme, but follow the same tendencies as predicted by the models.

\paragraph{Prediction peaks indicate lack of number understanding.} From earlier studies we know that BERT-based models have difficulties dealing with numbers in text \citep{bert-numbers}. The peaks that we observe in some predictions support this finding. For instance, the models assign a higher risk of \textit{Cardiac dysrhythmias} to patients aged 73 than to patients aged 74, because they do not capture that these are consecutive ages. Therefore, the influence of age on the predictions is solely based on the individual age tokens observed in the training data.

\subsection{Influence of Ethnicity} 

\paragraph{Mention of any ethnicity decreases prediction of mortality risk.} Table \ref{table:mortality-eth} shows the mortality predictions when different ethnicities are mentioned and when there is no mention. We observe that the mention of any of the ethnicities leads to a decrease in mortality risk prediction in all models, with White and African American patients receiving the lowest probabilities.

\begin{figure*}[t!]
  \includegraphics[width=\textwidth]{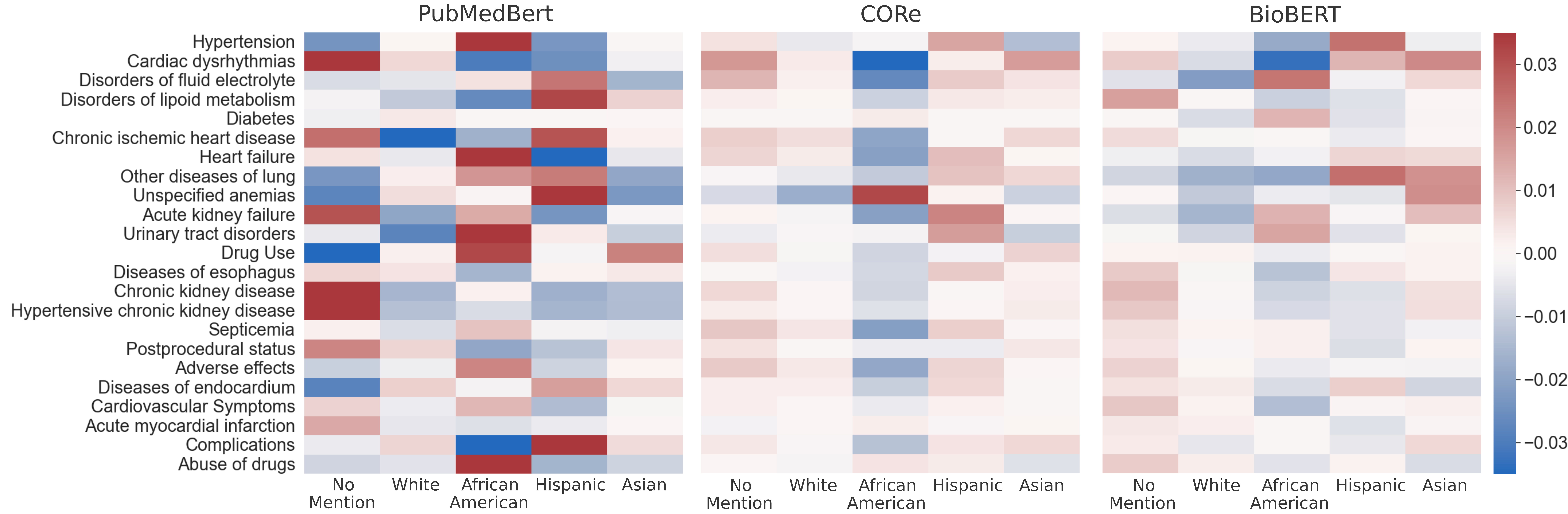}
    \caption{Influence of \textbf{ethnicity} on diagnosis predictions. Blue: Predicted probability for diagnosis is below-average; red: predicted probability above-average. PubMedBERT's predictions are highly influenced by ethnicity mentions, while CORe and BioBERT show smaller deviations, but also disparities on specific groups.}
\label{fig:ethnicity}
\end{figure*}

\begin{figure}[t!]
  \includegraphics[width=0.48\textwidth]{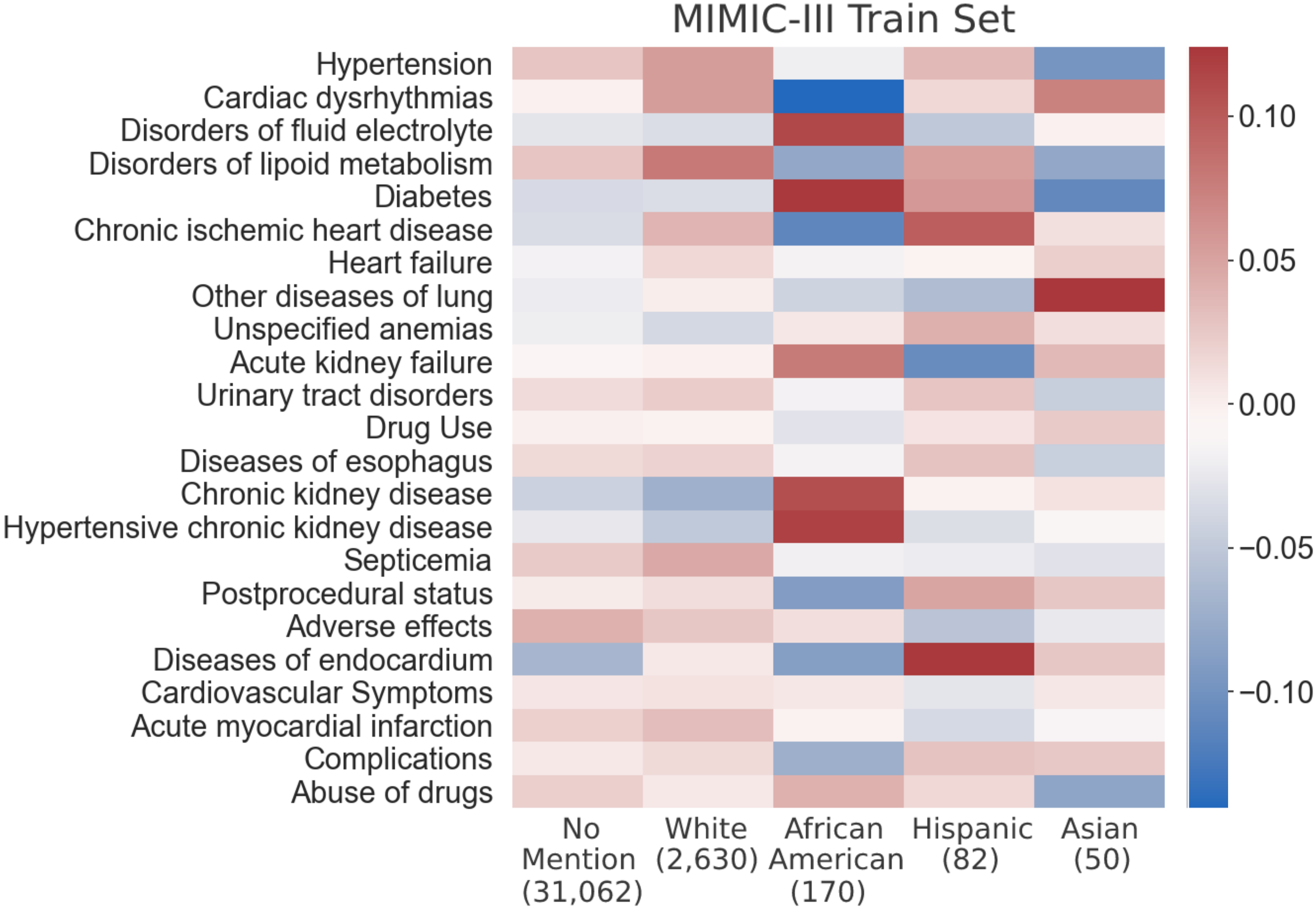}
  \caption{Original distribution of diagnoses per \hbox{\textbf{ethnicity}} in MIMIC-III. Cell colors: Deviation from average probability. Numbers in parenthesis: Occurrences in the training set. Both the distribution of samples and the occurrences of diagnoses are highly unbalanced in the training set. Some patterns are inherited by the fine-tuned models, while others are not.}
\label{fig:ethnicity-mimic}
\end{figure}

\paragraph{Diagnoses predicted by PubMedBERT are highly sensitive to ethnicity mentions.} Figure \ref{fig:ethnicity} depicts the influence of ethnicity mentions on the three models. Notably, the predictions of PubMedBERT are strongly influenced by ethnicity mentions. Multiple diagnoses such as \textit{Chronic kidney disease} are more often predicted when there is no mention of ethnicity, while diagnoses like \textit{Hypertension} and \textit{Abuse of drugs} are regarded more likely in African American patients and \textit{Unspecified anemias} in Hispanic patients. While the original training data in Figure \ref{fig:ethnicity-mimic} shows the same strong variance among ethnicities, this bias is not inherited the same way in the CORe and BioBERT models. However, we can also observe deviations regarding ethnicity in these models.

\begin{table}[b!]
  \caption{Influence of \textbf{ethnicity} on mortality predictions. The mention of an ethnicity decreases the predicted mortality risk. White and African American patients are assigned with the lowest mortality risk (gray-shaded).}
  \label{table:mortality-eth}
  \centering
  \begin{tabular}{lcccl}
    \toprule
    & PubMedBERT & CORe & BioBERT & \\
    \midrule
    No mention & \textbf{0.333} & \textbf{0.243} & \textbf{0.120} &  \\ 
    White & \cellcolor{gray!30} 0.329 & \cellcolor{gray!30} 0.235 & 0.119 &  \\ 
    African Amer. & \cellcolor{gray!30} 0.329 & 0.239 & \cellcolor{gray!30} 0.116 &  \\ 
    Hispanic & 0.331 & 0.237 & 0.118 &  \\ 
    Asian & 0.330 & 0.238 & 0.118 &  \\ 
    \bottomrule
  \end{tabular}
\end{table}

\paragraph{African American patients are assigned lower risk of diagnoses by CORe and BioBERT.} The heatmaps showing predictions of CORe and BioBERT reveal a potentially harmful pattern in which the mention of \textit{African American} in a clinical note decreases the predictions for a large number of diagnoses. This pattern is found more prominently in the CORe model, but also in BioBERT. This behavior can lead to disadvantages in the treatment of African American patients and would reinforce existing biases in health care \citep{race-treatment}.

\section{Discussion}
\paragraph{Sensitivity and impact of characteristics show large variance.} The results described in \ref{section:results} reveal large differences in the influence of patient characteristics throughout models. The analysis shows that there is no overall \textit{best} model, but each model has learned both useful patterns (e.g. age as a medical plausible risk factor) and potentially dangerous ones (e.g. decreases in diagnosis risks for minority groups). The large variance is surprising since the models have a shared architecture and are fine-tuned on the same data--they only differ in their pre-training. And while the reported AUROC scores for the models (Table \ref{table:models}) are close to each other, the variance in learned behavior show that we should consider in-depth analyses a crucial part of model evaluation in the clinical domain. This is especially important since unintended biases in clinical NLP models are often fine-grained and difficult to detect.

\paragraph{Best performing model is especially sensitive to gender and ethnicity mentions.} The analysis has shown that PubMedBERT which outperforms the other models in both mortality and diagnosis prediction show larger sensitivity to mentions of gender and ethnicity in the text. This is alerting since it particularly affects  minority groups which are already disadvantaged by the health care system. It also shows that instead of measuring clinical models regarding a single score, looking at their robustness and potential impact should be further emphasized.

\paragraph{De-biasing methods need to be aligned with medical knowledge.} The application of de-biasing approaches has shown to be effective in general language scenarios in the past \citep{debias}. While their evaluation is out of the scope of this work, we want to highlight that their application in clinical outcome prediction can be challenging. We argue that de-biasing methods cannot be applied to patient characteristics in clinical text in the same way as for general language. The decision about which characteristics should be considered a risk factor and their impact on outcome predictions should be aligned with medical knowledge. Therefore, we focus followup research towards iterative model learning using feedback loops with medical professionals to define favorable patterns and adverse ones.





\section{Conclusion}
In this work, we introduced a novel behavioral testing framework for the clinical domain that enables us to understand the effects of textual variations on the model's prediction. We apply this framework to examine the impact of certain patient characteristics, and evaluate whether current NLP models reproduce dangerous biases in health care. Our results show that the models have indeed learned to overestimate certain characteristics especially those of minority groups which potentially lead to disadvantages. With this work we want to emphasize the importance of model evaluation beyond common metrics especially in sensitive areas like health care. For future research we propose additional behavioral analyses, e.g. regarding stigmatizing language in clinical notes as defined by \citet{words-matter}. We also propose to apply the framework to evaluate different de-biasing approaches and to further develop approaches for removing harmful biases while keeping plausible patterns regarding clinical risk factors intact.

\section*{Acknowledgments}
Our work is funded by the German Federal Ministry for Economic Affairs and Energy (BMWi) under grant agreement 01MD19003B (PLASS) and 01MK2008MD (Servicemeister).

\bibliography{custom}

\end{document}